\documentclass[10pt,twocolumn,letterpaper]{article}

\usepackage[pagenumbers]{wacv} % To force page numbers, e.g. for an arXiv version

\usepackage{graphicx}
\usepackage{amsmath}
\usepackage{amssymb}
\usepackage{booktabs}

\usepackage{multirow}
\usepackage{graphicx}
\usepackage{float}
\usepackage[normalem]{ulem}
\useunder{\uline}{\ul}{}

\usepackage{enumitem}
\usepackage[accsupp]{axessibility} % Improves PDF readability for those with disabilities.

\usepackage[pagebackref,breaklinks,colorlinks]{hyperref}

\usepackage[capitalize]{cleveref}
\crefname{section}{Sec.}{Secs.}
\Crefname{section}{Section}{Sections}
\Crefname{table}{Table}{Tables}
\crefname{table}{Tab.}{Tabs.}

\def\wacvPaperID{1305} % *** Enter the WACV Paper ID here
\def\confName{WACV}
\def\confYear{2024}

\begin{document}

\title{C-CLIP: Contrastive Image-Text Encoders to Close the Descriptive-Commentative Gap}

\author{William Theisen\\
University of Notre Dame\\
{\tt\small wtheisen@nd.edu}
\and
Walter Scheirer\\
University of Notre Dame\\
}
\maketitle

%%%%%%%%% ABSTRACT
\begin{abstract}
The interplay between the image and comment on a social media post is one of high importance for understanding its overall message.
Recent strides in multimodal embedding models, namely CLIP, have provided an avenue forward in relating image and text.
However the current training regime for CLIP models is insufficient for matching content found on social media, regardless of site or language. Current CLIP training data
is based on what we call ``descriptive'' text: text in which an image is merely described. This is something rarely seen on social
media, where the vast majority of text content is ``commentative'' in nature. The captions provide  commentary and broader context related to the image, rather than describing what is in it. Current CLIP models perform poorly on retrieval tasks where image-caption pairs display a commentative relationship. Closing this gap would be beneficial for several important application areas related to social media. For instance, it would allow groups focused on Open-Source Intelligence Operations (OSINT) to further aid efforts during disaster events, such as the ongoing Russian invasion of Ukraine, by easily exposing data to non-technical users for discovery and analysis. In order to close this gap we demonstrate that training contrastive image-text encoders on explicitly commentative pairs results in  large improvements in retrieval results, with the results extending across a variety of non-English languages.
\end{abstract}

%%%%%%%%% BODY TEXT
\section{Introduction}
\label{sec:intro}

\begin{figure}[!t]
\centering
\includegraphics[width=1.0\linewidth]{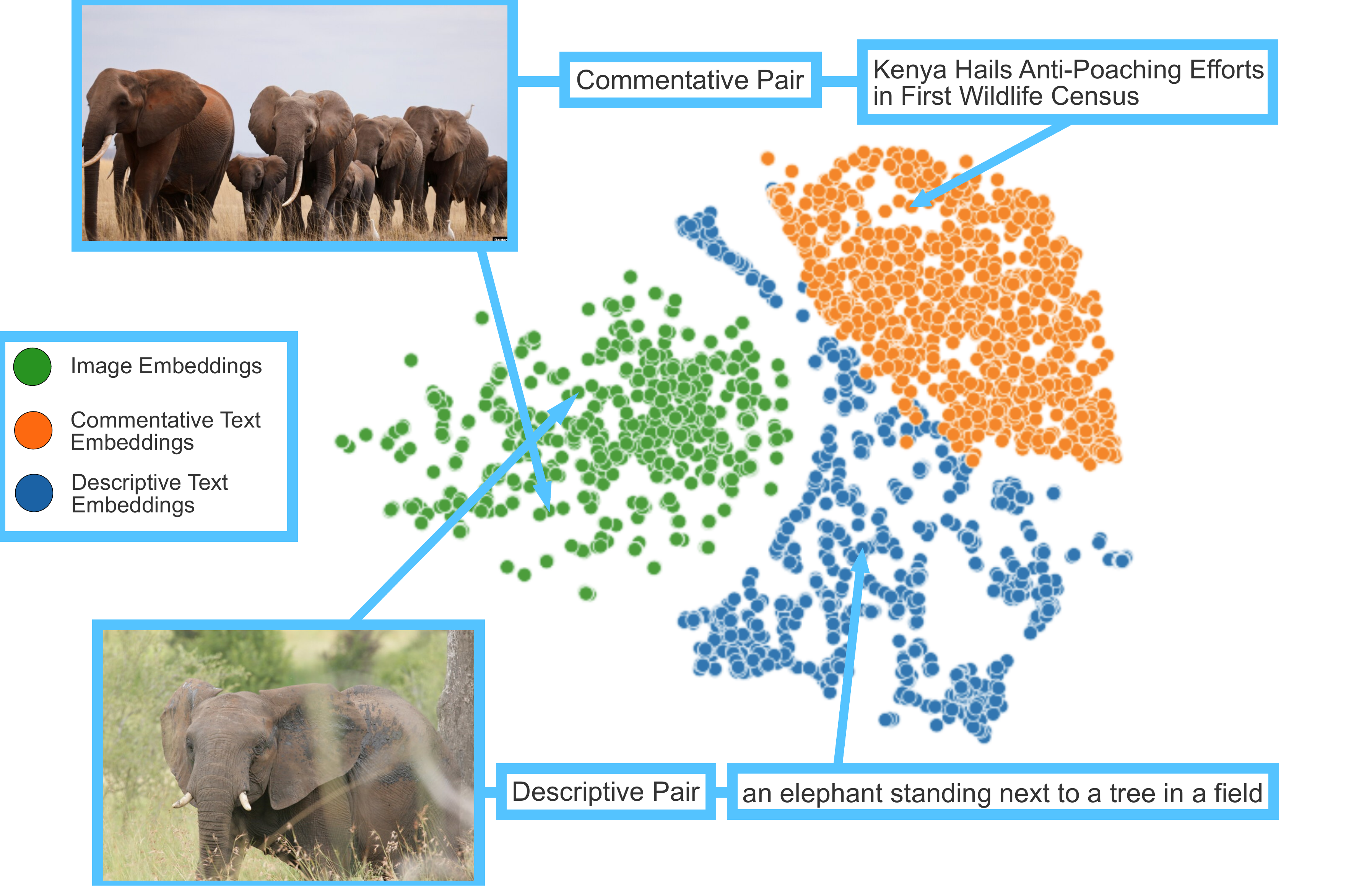} % Reduce the figure size so that it is slightly narrower than the column.
% \vspace{-2mm}
\caption{A TSNE~\cite{JMLR:v9:vandermaaten08a} reduction of text and image embeddings from a baseline CLIP model. The two groups of texts are taken from the MSCOCO~\cite{coco} data set commonly used to train CLIP models and taken from a data set of social media posts on the website Telegram~\cite{theisen}. While the images from the two groups share an area of the latent space (green) the text embeddings map to two distinct regions, showing the fundamental difference between commentative text (orange) and descriptive text (blue), which we term the description-commentary gap.}
\vspace{-2mm}
\label{fig:tease}
\end{figure}

\begin{figure*}[t!]
\centering
\includegraphics[width=0.98\textwidth]{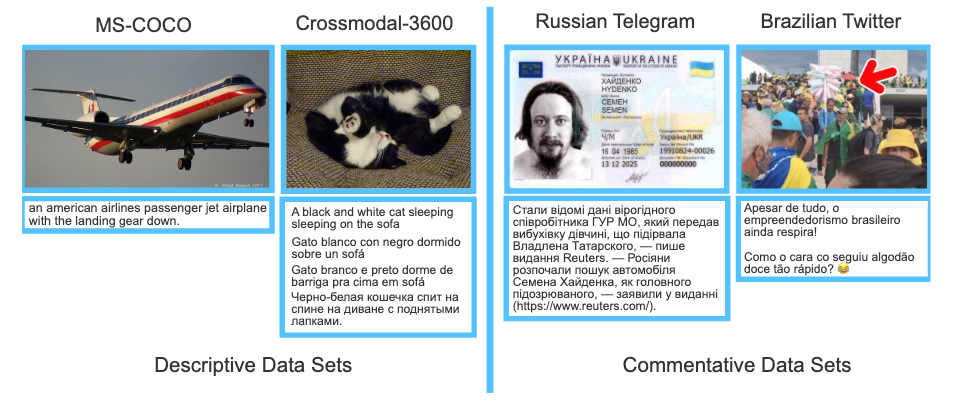} % Reduce the figure size so that it is slightly narrower than the column.
% \vspace{-2mm}
\caption{Example images and their caption(s) from each of the four data sets. The four data sets are split into two groups, descriptive (left) and commentative (right). There are five languages reflected in these data sets: English, Spanish, Portuguese, Ukrainian, and Russian.}
\vspace{-2mm}
\label{fig:data}
\end{figure*}

Current publicly available Constrastive Language-Image Pre-Training (CLIP)~\cite{radford2021learning} models suffer from a hidden problem which we term the ``Descriptive-Commentative Gap'' (hereafter DCG). CLIP models are trained on text we define as descriptive, the text is "presenting observations about the characteristics of someone or something", with the thing in this case being the image it is paired with. If the picture is an elephant, the text will simply say "an elephant standing next to a tree in a field", it provides no additional information or context beyond what the objects in the image represent. However, image-text pairs encountered in everyday life are rarely of a descriptive relationship. Social media image-text pairs are often of a commentative nature, with the text giving "an expression of opinions or offering of explanations about an event or situation" while the image is of the event or situation at hand. Given another photo of an elephant, a simple commentative text pairing could be "Kenya Hails Anti-Poaching Efforts in First Wildlife Census"~\cite{voanewsKenyaHails}. The image itself is not directly reflected in the text. CLIP models often report accuracies of up to 99\% on retrieval tasks where the text is description. When tested on posts where the text is commentary rather than description the accuracy drops by 60\% on average. The transferability of the models to the ``wild-west'' of commentative captioning is poor, thus lowering the models abilities to be used by others for downstream tasks.

Open-Source intelligence has gained an increasing amount of publicity as the amount of data available has continued to grow larger. Current estimates value the OSINT market at \$6.25B and with a Compound Annual Growth Rate (CAGR) of 25 percent, valuations in 2033 reach \$58.21B~\cite{osint}. 
%Using publicly available information, such as social media posts, policy makers, journalists, and academics are able to better understand complex events and subsequently make better decisions in some cases. 
The invasion of Ukraine has provided many recent examples of OSINT in action, The Economist provides several examples; during the counter-offensive ``amateur analysts on Twitter tracked the Ukrainian advance, almost in real time, by `geo-locating' the images contrasted with a Russian soldier posting pictures from the front-lines.'' ``His post included a geo-tag of the exact location. Ukrainian missiles later struck it''~\cite{The_Economist_2023}. Of vital importance to OSINT operators is the ability to sift through massive amounts of data. Much of this work is done manually, but with the recent release of high quality multimodal models such as CLIP ~\cite{radford2021learning}, there is hope that much of this work can be automated in the near-future, as unfortunately the DCG prevents us from doing so now.

In Fig.~\ref{fig:tease} one can see a visualization of the DCG. In orange are the commentative captions from a data set of Russian Telegram posts~\cite{theisen}. In blue are the descriptive captions from the MSCOCO~\cite{coco} dataset. The two groups of texts are contained in two clearly demarcated regions of the latent space, while the images (in green) share a grouping in the space. Due to the differences in style between commentary and description, the embeddings of the two classes of text are different, leading to decreased accuracy on downstream tasks when the task is operating on commentary style data. Little discussion of the DCG has been had in technical literature to date. 
 To help with this problem the paper makes the following contributions:
\begin{enumerate}[noitemsep,nolistsep]
    \item Defines and quantifies the Description-Commentary Gap.
    \item Quantifies the gap across 5 languages and 3 social media sites.
    \item Proposes solutions for closing the gap.
    \item Lists experiments for testing models resulting in...
    \item Several newly trained models achieving state-of-the-art retrieval accuracies on social media related downstream tasks, open-sourced on Huggingface~\cite{wolf2020huggingfaces}.  
\end{enumerate}

\begin{figure*}[t!]
\centering
\includegraphics[width=0.98\textwidth]{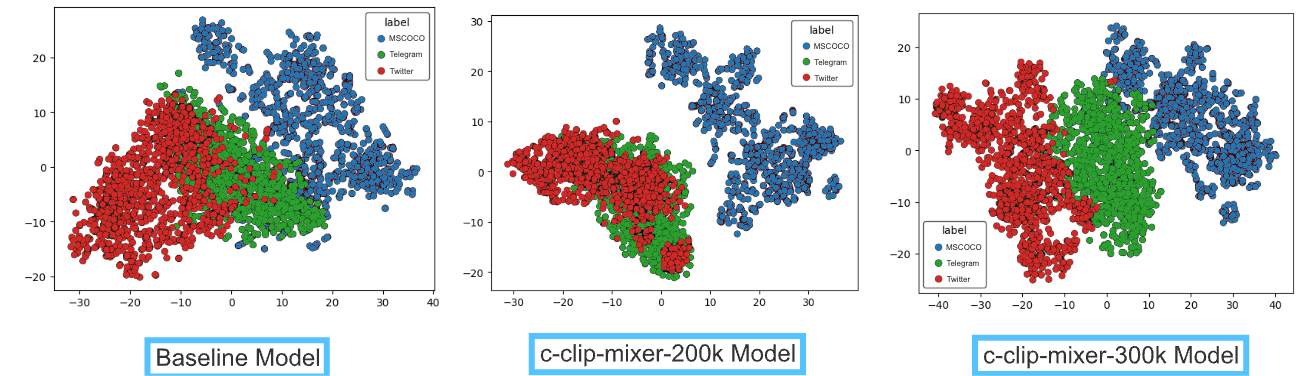} % Reduce the figure size so that it is slightly narrower than the column.
% \vspace{-2mm}
\caption{TSNE plots for the latent spaces of 3 models. On the left is a baseline model, showing a clear separation between the two commentative data sets and the descriptive data. In the middle can be seen what happens when the projection layers are trained on a combined set of Twitter and Telegram commentative data, a convergence of the two groups happens, which is expected. Surprisingly, when a model is trained on a mixed set of commentative and descriptive data a convergence is not seen, and instead order is applied to the three groups but they are kept essentially distinct.}
\vspace{-2mm}
\label{fig:dcg}
\end{figure*}

\section{Related Work}

%The description-commentary gap has yet to be formally discussed in the literature. 
Several papers have raised the issue that not all social media comments are descriptions of their accompanying image, but aside from recognizing this issue it appears an open problem. Vempala and Preotiuc-Pietro claim that " little is known about how textual content is related to the images with which they appear" and describe a grid by which they go on to categorize text-image relations~\cite{vempala-preotiuc-pietro-2019-categorizing}. An image may either add to a tweets meaning or not, and text may be represented in an image or not. 
%However with this data set they make no attempt to solve the problem, instead training a classifier to try and predict which of these 4 categories a tweet could fall into. 
Sosea et al. also recognize that tweets and their images may not be of a descriptive character, with their categories consisting of: unrelated, similar, and complementary~\cite{tweet_disaster}. While both works recognize that this issue exists they make no attempt to formalize it nor theorize that current model training regimes fail to capture this difference. Additionally both works were published prior to the publication of CLIP and therefore do not and cannot, explore the possibilities this family of encoders provides.

The publication of the Contrastive Language-Image Pre-Training (CLIP) by Radford et al. was a leap forward in multimodal modeling~\cite{radford2021learning}. By jointly training the image and text encoders together on the pairs they easily achieved state-of-the-art performance on downstream tasks such as image retrieval. However, many of these initial models were closed-source and were entirely focused on English. To allay these concerns Carlsson et al. published several multilingual CLIP models, including one which they claim beats the original English-only CLIP model on several benchmarks~\cite{carlsson-EtAl:2022:LREC}. Nils Riemers and Iryna Gurevych also published a multilingual model, using siamese bert networks~\cite{reimers-2019-sentence-bert}.

The models used in this paper are merely built on top of pre-existing work. For the vision model we focus on transformers primarily as introduced by Dosovitskiy et al.~\cite{dosovitskiy2021image}. More specifically we use the original clip-vit-base-patch32 model as released by openai~\cite{radford2021learning}. We pair with this a variety of Bert models, originally introduced by Devlin et al.~\cite{devlin2019bert}. The primary focus was on RoBerta models~\cite{conneau2020unsupervised} and distilBert models~\cite{sanh2020distilbert}. The highest accuracy was achieved using a distlBert model that had been trained with multilingual knowledge distillation as introduced by Nils Reimers and Iryna Gurevych~\cite{reimers2020making}.

Downstream multilingual tasks have been discussed in works such as "Towards Zero-shot Cross-lingual Image Retrieval" by Aggarwal and Kale~\cite{aggarwal2021zeroshot} and by Nascimento et al~\cite{nascimento2022fewshot}. Aggarwal and Kale discuss training regimes for expanding the models beyond English using pre-training on the text encoder, but as their data set is primarily an extension of MS-COCO~\cite{coco}, they focus only on descriptive pairs. Nascimento et al. focuses on social media data, Brazilian tweets surrounding a series of protests, but they focus on data curation to attempt to reduce the need for labelled data when training GNNs.

The description-commentary gap rears its head primarily when a CLIP model is applied to data collected from social media. ~\cite{aggarwal2021zeroshot} and ~\cite{nascimento2022fewshot} both focus on tweets, but social media at large, and especially memes, have taken a much larger seat at the table of online content understanding.

When considering social media data, image captioning is a closely related downstream task. Coming to the forefront in 2015 with the papers by Xu et al.~\cite{DBLP:journals/corr/XuBKCCSZB15} and Vinyals et al.~\cite{DBLP:journals/corr/VinyalsTBE14}, and more recently Cornia et al.~\cite{Cornia_2019_CVPR}; the goal of image captioning is, given an image, generate caption for this image. MSCOCO features heavily in these papers, and as a result the captions generated are of a descriptive nature. By utilizing object detection frameworks, the papers all simply focus on the recognized objects in the image, yielding captions such as "a little girl in a pink hat is blowing bubbles." More recently, several papers have begun focusing on image captioning specifically as it relates to social media. Wang et al.~\cite{Wang2018-ww} state that "the social image, associated with a set of user-contributed tags, has been rarely investigated for a similar task." Unfotunately they again simply fall into taking objects as "tags" for their tasks, resulting in image objects described in text as the returned captions. We maintain that the average post on social media does not take this form of image-caption relationship, and CLIP models provide a road forward in improving this modeling.

Further downstream tasks such as retrieval and clustering for social media has also been well treated in prior literature. Zannettou et al.~\cite{zanne}, Dubey et al.~\cite{dubey} both cover the clustering of image macros. Unfortunately they fail to bridge the gap of multi-modality, with both raising it as an area of future work. Beskow et al.~\cite{beskow} attempts to bridge the gap between multiple modalities using deep-neural networks but  restricts their definition of meme to a picture with superimposed white text in impact font and/or text placed in a white space over a picture. The three previous approaches are all supervised methods requiring strict categorization of the image content, Theisen et al.~\cite{theisen} states that an unsupervised approach yields better results when treating with social media data as virtually no categories of images can be assumed and new categories are constantly popping up, especially surrounding new events. It is with this in mind that we propose training on large unsorted collections of social media image-pairs in order to understand and close the descriptive-commentary gap.

\section{Methodology}

\textbf{Descriptive-Commentative Gap:} We hypothesize that there is a quantifiable difference between text used for training CLIP models and text that would appear in downstream tasks relating to social media. We call this difference the "Description-Commentary Gap", hereafter DCG. In Figure 2 we can see two examples of image-text pairs, one taken from the COCO data set frequently used to train CLIP models and another taken from the social media site Telegram. The left pair, from COCO, is what could be described as a "descriptive" pair. The text simply describes or lists the objects that appear in the image. Contrasted to this is the right pair with the text being, at the surface level, almost completely disconnected from the image. They share no objects. This text is of a "commentative" nature with the text adding additionally context to the image and relating it in some way that is not immediately obvious. These two pairs illustrate the DCG.

We hypothesize that the existence of the DCG is non-trivial and significantly reduces accuracies on downstream tasks involving social media data, which is almost entirely of a commentative nature. Humans can intuit this gap, but it has yet to be actually quantified in prior literature and is therefore worth exploring further. Figure 1 shows that the DCG also appears in the latent space of CLIP text embeddings, spanning across a number of different possible text models, languages, and datasets. Descrpitive embeddings are a discrete group showing a degree of separation from commentative embeddings.

Theisen et al. ends their paper with the belief "that using all available context is of the
utmost importance for future studies in this area." The release of CLIP models provides an avenue forward in connecting the multimodal aspects of social media posts. However the DCG is a yet unanswered question in the way, reducing accuracy on tasks relating to social media data and preventing non-technical works from benefiting from this technology. To help alleviate this problem a number of different, publicly available, models were trained which we hope can be of use to those wishing to model social media content multimodally.

%\section{Methodology}
\begin{table}[]
\centering
\begin{tabular}{l|l}
Training Argument & Value \\ \hline
Epochs            & 50    \\
Early Stopping    & True  \\
Batch Size        & 32    \\
Learning Rate     & 5e-5  \\
Optimizer         & Adam~\cite{adam}  \\
Max Seq. Length   & 128  
\end{tabular}
\caption{Basic training parameters used in all C-CLIP models.}
\label{tab:trainer}
\end{table}

\textbf{Data and Procedures:} In order to establish that the DCG exists, we test three baseline models. OpenAI's publicly available CLIP model~\cite{radford2021learning}, a multilingual model from the M-CLIP team~\cite{carlsson-EtAl:2022:LREC}, and a multilingual model from the SentenceTransformers group~\cite{reimers-2019-sentence-bert}. For this task we use four data sets, two descriptive data sets, one in English and a multilingual set and two commentative data sets, in languages included in the multilingual models training data. We include three non-english language data sets to demonstrate that the DCG is language-agnostic. The two descriptive data sets are Microsoft's Common Objects in COntext (COCO)~\cite{lin2015microsoft}, a standard image-caption data set and Cross-modal 3600 (XM3600)~\cite{thapliyal2022crossmodal3600}, a data set from google released specifically to test multilingual multimodal models in a variety of languages. For our task we choose captions from five languages: English, Spanish, Ukrainian, Russian, and Portuguese. 

The two commentative data sets are tweets surrounding a series of protests in Brazil~\cite{nascimento2022fewshot} and Telegram posts related to the ongoing war in Ukraine~\cite{theisen}. Figure~\ref{fig:data} shows several examples from both data sets and hopefully allows a reader to intuit the fundamental difference between the caption-image pairings across the two categories. These data sets are chosen for a variety of reasons. In addition to their essentially commentative captioning, being directly scraped from social media, they both span a variety of languages. Additionally, they both cover events that have been of particular interest in the OSINT space. According to the Verge, "Telegram has become a window into the war." and while not a common source of social media in western countries "in Russia, Telegram has become sometimes the only source of information amid stifling government censorship. Across the border, the platform has become a lifeline for Ukrainians trying to keep safe from Russia’s attacks and track troop movements. And for the rest of the world, Telegram has become the window into a war that has destabilized the world."~\cite{Borak_2023} The January 8th protests that rocked Brazil were another flashpoint for OSINT operators, with people working on twitter data to help authorities track down and unmask protesters~\cite{Maverick_2023}. We hope that by training models on data sets of already high importance in the OSINT community we can immediately begin helping these efforts.

The data sets are also large enough to allow for the training of multiple models and a full suite of testing to be run. The Brazilian Twitter data set has 203,781 image-text pairs. The Russian Telegram data set is much larger, yielding 4,766,631 image-text pairs.

\begin{figure*}[t!]
\centering
\includegraphics[width=0.98\textwidth]{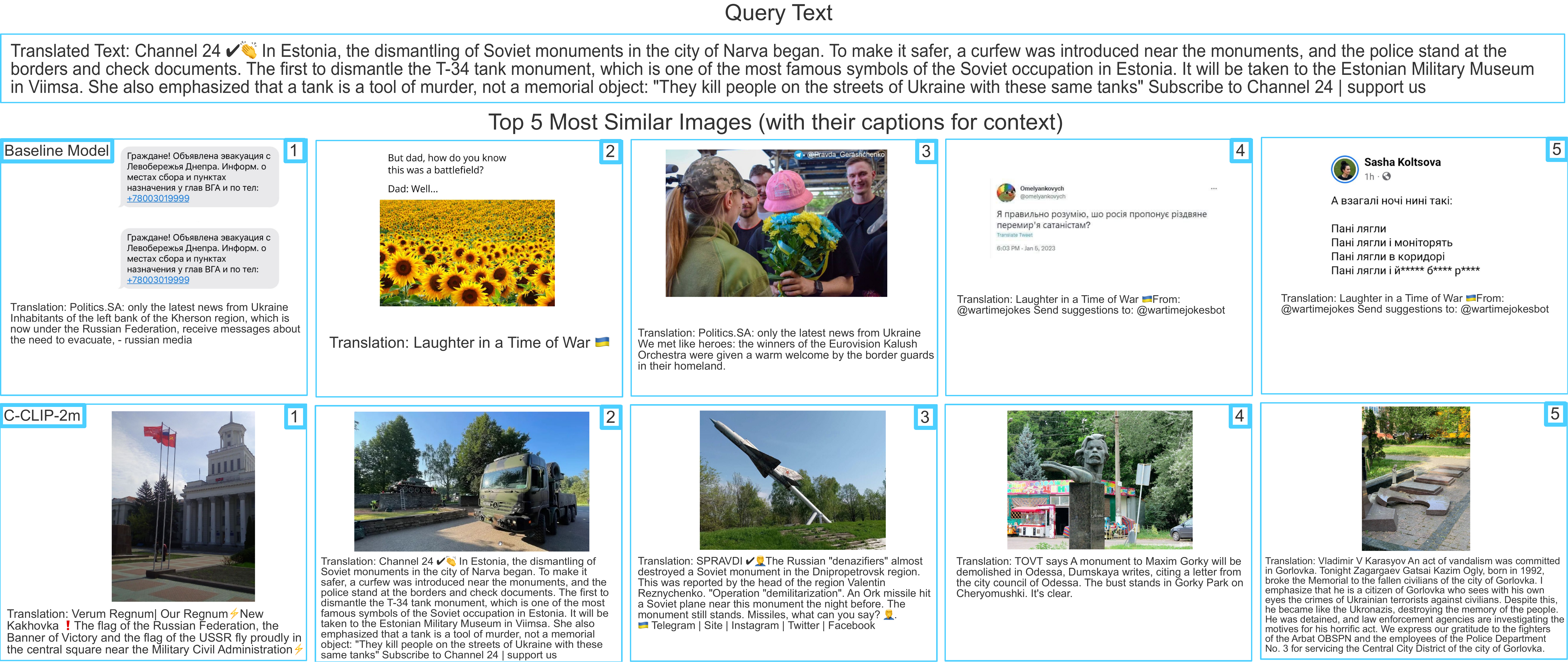} % Reduce the figure size so that it is slightly narrower than the column.
% \vspace{-2mm}
\caption{The top 5 most similar images as measured by two CLIP models from a population of 1000 image-text pairs for a query text. The baseline model returns only spurious results, with no clear connection between the query text (top) and the returned images (middle). However the C-CLIP-2m model (bottom) returns 4 images of monuments and 1 image of a soviet symbol (the flag), both of which are mentioned in the first query text. Qualitatively it can be seen that C-CLIP-2m learns quite well the relations between commentative text and their images. We provide also the translated captions that accompanied each image, to give context for the reader. The original, untranslated text, is available in the supplementary material.}
\vspace{-2mm}
\label{fig:quali}
\end{figure*}

After demonstrating the existence of the DCG qualitatively and quantitatevly, we trained a number of models with the intent to improve performance of CLIP models on tasks using commentative data. Using the VisionTextDualEncoder (VTDE)~\cite{vtde} module from HuggingFace one can easily pair pre-trained image and text models. 

The vision model is the same used in prior works~\cite{radford2021learning}~\cite{carlsson-EtAl:2022:LREC}~\cite{reimers-2019-sentence-bert}, being a ViT with an output dimension of 768. This is paired with the DistilBert model from Riemers et al.~\cite{reimers-2019-sentence-bert}, also with an output dimension of 768. The VDTE forward function had to be modified to correctly use the DistilBert model. Instead of directly using the built in text\_projection function, the last hidden state had to be averaged over the number of the tokens in the input string prior to the call. These were then used to calculate the loss. Via the two projection layers added on top of the dual models the output dimensions were projected into the shared latent space and reduced from the original dimension of 768 to 512.

The VTDE then places on top of the two models two projection layers, which then must be further trained on a downstream task.

To train the models, we use HuggingFace's Trainer class~\cite{trainer} with the arguments outlined in Table~\ref{tab:trainer}. These settings are rather boilerplate and were based off the work done in Bianchi et al.~\cite{bianchi2021contrastive} where the Italian authors fine-tuned a CLIP model to extend to their language. and the Github repository~\cite{github}.

\begin{table*}[]
\centering
\resizebox{\textwidth}{!}{%
\begin{tabular}{ccccc|llll|llll|llll}
\multirow{2}{*}{\begin{tabular}[c]{@{}c@{}}Model Retrieval\\ Accuracy (all-pairs)\end{tabular}} & \multicolumn{4}{c|}{MS-COCO} & \multicolumn{4}{c|}{XM3600} & \multicolumn{4}{c|}{Russian Telegram} & \multicolumn{4}{c}{Brazilian Tweets} \\ \cline{2-17} 
 & @1 & @5 & @10 & @25 & \multicolumn{1}{c}{@1} & \multicolumn{1}{c}{@5} & \multicolumn{1}{c}{@10} & \multicolumn{1}{c|}{@25} & \multicolumn{1}{c}{@1} & \multicolumn{1}{c}{@5} & \multicolumn{1}{c}{@10} & \multicolumn{1}{c|}{@25} & \multicolumn{1}{c}{@1} & \multicolumn{1}{c}{@5} & \multicolumn{1}{c}{@10} & \multicolumn{1}{c}{@25} \\ \hline
OpenAI CLIP~\cite{radford2021learning} & 57.3\% & 96\% & 99.2\% & 100\% & - & - & - & - & - & - & - & - & - & - & - & - \\
Sen.-Tran. CLIP~\cite{reimers-2019-sentence-bert} & 65.89\% & 89.1\% & 96.1\% & 99.70\% & 41.06\% & 69.13\% & 78.86\% & 88.73\% & 6.06\% & 13.18\% & 17.08\% & 24.08\% & 10.40\% & 19.53\% & 24.55\% & 32.79\% \\
M-CLIP~\cite{carlsson-EtAl:2022:LREC} & 64.93\% & 74.44\% & 83.69\% & 92.19\% & 72.6\% & 92.89\% & 96.36\% & 98.87\% & 12.99\% & 24.55\% & 30.75\% & 39.80\% & 5.71\% & 11.78\% & 15.32\% & 21.28\% \\ \hline
C-CLIP-2M & 4.4\% & 7.7\% & 13.5\% & 24.5\% & 3.8\% & 13.27\% & 19.99\% & 34.9\% & \textbf{29.55\%} & \textbf{57.25\%} & \textbf{67.26\%} & \textbf{79.41\%} & 1.37\% & 4.93\% & 7.63\% & 13.8\%\\
C-CLIP-Mixer-200k & 3.6\% & 6.3\% & 9.2\% & 18.8\% & 3\% & 10.87\% & 17.49\% & 30.93\% & 7.34\% & 21.30\% & 29.14\% & 42.73\% & \textbf{13.69\%} & \textbf{34.28\%} & \textbf{45.35\%} & \textbf{60.99\%} \\
C-CLIP-Mixer-300k & 18.4\% & 24.8\% & 34.5\% & 51.9\% & 6.4\% & 18\% & 28.5\% & 45.6\% & 7.9\% & 18.7\% & 27.8\% & 37.7\% & 12.37\% & 30.83\% & 41.93\% & 56.87\% \\
%C-CLIP-mixed-Big & - & - & - & - & 3.13\% & 10.93\% & 17.03\% & 30.73\% & 26.19\% & 51.96\% & 63.22\% & 75.83\% & 9.30\% & 24.41\% & 33.35\% & 47.72\%
\end{tabular}%
}
\caption{Retrieval accuracy results for top-performing models across the four data sets, with a population size of 1000 image-text pairs. The three baseline models achieve fantastic results on the descriptive data sets of MS-COCO and XM3600, but dismal accuracies on the two commentative data sets. Contrasted to this are the two C-CLIP models getting the highest results on the two commentative data sets. It shows that the commentative image-text pair space can be learned, but does not extend back to the descriptive space. There appears to be a trade-off required between the two types of data sets. The bottom row is our attempt at training a model to perform equally well on all data sets.}
\label{tab:baselines}
\end{table*}

Several training regimes were tested. The first is training a single model on a single data set style, for example training a Russian Telegram model specifically on collections of Russian Telegram data of varying sizes. Additionally tested was mixing training data in both even and uneven splits, attempting to understand whether the DCG is truly cross-site and multilingual, and what the impacts on performance may be if a model was trained on a mixture of Brazilian Tweets and Russian Telegram, or Russian Telegram and descriptive data.

Models were trained at 10,000 image-text pairs and 100,000 image-text pairs. For Russian Telegram, due to the large size of the data set, 1 million and 2 million pairs were trained as well. In addition to these training splits, validation splits of 20,000 pairs were left out. A final set of 111,000 pairs was used for testing. 1,000 for 10 trials at a population of 100 pairs, 10,000 for 10 trials at 1,000 pairs, and 100,000 for 10 trials with 10,000 pairs. The largest model (C-CLIP-tele-2m) took approximately 28 hours to train. As a rough rule of thumb, every 10,000 pairs added 10 minutes to the training time (on a Titan X with 12GB of VRAM).

To establish the general performance of our models we then calculate the retrieval accuracy of each model. For population sizes of 100, 1000, and 10000 we compute the pairwise cosine similarity of all image text pairs. Then at recalls of 1, 5, 10, and 25 we compute the retrieval accuracy for each artifact. These accuracy results were averaged across ten trials for each population size. These were then compared against the three baseline models, tested in the same manner. One limitation of the original CLIP model is that it is only trained on English data and thus the performance on MSCOCO is only to establish that the multilingual baseline models achieve similiar accuracies and therefore allow for a reasonable point of comparison to our work. 

\section{Experiments and Results}
\label{sec:experiments}

\begin{table*}[]
\centering
\resizebox{\textwidth}{!}{%
\begin{tabular}{cllll|llll|llll}
\multirow{2}{*}{\begin{tabular}[c]{@{}c@{}}Telegram Retrieval\\ Accuracy (all-pairs)\end{tabular}} & \multicolumn{4}{c|}{Pop=100} & \multicolumn{4}{c|}{Pop=1,000} & \multicolumn{4}{c}{Pop=10,000} \\ \cline{2-13} 
 & \multicolumn{1}{c}{@1} & \multicolumn{1}{c}{@5} & \multicolumn{1}{c}{@10} & \multicolumn{1}{c|}{@25} & \multicolumn{1}{c}{@1} & \multicolumn{1}{c}{@5} & \multicolumn{1}{c}{@10} & \multicolumn{1}{c|}{@25} & \multicolumn{1}{c}{@1} & \multicolumn{1}{c}{@5} & \multicolumn{1}{c}{@10} & \multicolumn{1}{c}{@25} \\ \hline
M-CLIP (Baseline)~\cite{carlsson-EtAl:2022:LREC} & 23.40\% & 38.80\% & 47.70\% & 66.70\% & 6.06\% & 13.18\% & 17.08\% & 24.08\% & 2.88\% & 5.85\% & 7.82\% & 11.02\% \\
C-CLIP-tele-2m & {\ul \textbf{54.50\%}} & {\ul \textbf{81.30\%}} & {\ul \textbf{88\%}} & {\ul \textbf{96.29\%}} & {\ul \textbf{29.55\%}} & {\ul \textbf{57.25\%}} & {\ul \textbf{67.26\%}} & {\ul \textbf{79.41\%}} & {\ul \textbf{10.25\%}} & {\ul \textbf{26.87\%}} & {\ul \textbf{36.59\%}} & {\ul \textbf{51.11\%}} \\
C-CLIP-tele-1m & 39.20\%& 63.40\% & 73.30\% & 86.10\% & \multicolumn{1}{c}{23.30\%} & \multicolumn{1}{c}{46.47\%} & \multicolumn{1}{c}{56.14\%} & \multicolumn{1}{c|}{67.90\%} & 8.02\% & 21.17\% & 29.22\% & 41.48\% \\
C-CLIP-tele-100k & 28.90\% & 51.10\%& 63.40\% & 79.20\% & 13.40\% & 30.60\% & 40.05\% & 53.55\% & 3.32\% & 10.40\% & 15.50\% & 24.72\% \\
C-CLIP-tele-10k & 16.09\% & 34.89\% & 48.59\% & 68.10\% & 4.60\% & 14.29\% & 21.29\% & 33.72\% & 0.86\% & 3.17\% & 5.33\% & 10.12\% \\ \\
\multirow{2}{*}{\begin{tabular}[c]{@{}c@{}}Twitter Retrieval\\ Accuracy (all-pairs)\end{tabular}} & \multicolumn{4}{c|}{Pop=100} & \multicolumn{4}{c|}{Pop=1,000} & \multicolumn{4}{c}{Pop=10,000} \\ \cline{2-13} 
 & \multicolumn{1}{c}{@1} & \multicolumn{1}{c}{@5} & \multicolumn{1}{c}{@10} & \multicolumn{1}{c|}{@25} & \multicolumn{1}{c}{@1} & \multicolumn{1}{c}{@5} & \multicolumn{1}{c}{@10} & \multicolumn{1}{c|}{@25} & \multicolumn{1}{c}{@1} & \multicolumn{1}{c}{@5} & \multicolumn{1}{c}{@10} & \multicolumn{1}{c}{@25} \\ \hline
M-CLIP (Baseline)~\cite{carlsson-EtAl:2022:LREC} & 18.9\% & 38.6\% & 48.2\% & 68.9\% & 5.71\% & 11.78\% & 15.32\% & 21.28\% & 4.11\% & 7.58\% & 9.9\% & 13.39\% \\
C-CLIP-twit-100k & 28\% & 59.9\% & 73.5\% & 90.1\% & 7.27\% & 21.47\% & 31\% & 46.7\% & 1.14\% & 3.93\% & 6.35\% & 11.39\%\\
C-Clip-twit-10k & 14.40\% & 38.5\% & 53.5\% & 76.1\% & \multicolumn{1}{c}{4.73\%} & \multicolumn{1}{c}{11.87\%} & \multicolumn{1}{c}{16.4\%} & \multicolumn{1}{c|}{27.97\%} & 0.84\% & 2.57\% & 4.05\% & 6.88\%\\
C-CLIP-Mixer-300k & 38.09\% & 69.4\% & 82.19\% & 92.9\% & 12.37\% & 30.83\% & 41.93\% & 56.87\% & 2.15\% & 6.77\% & 10.15\% & 16.67\% \\
C-CLIP-Mixer-200k & {\ul \textbf{39.4\%}} & {\ul \textbf{71.7\%}} & {\ul \textbf{82.7\%}} & {\ul \textbf{93.5\%}} &  {\ul \textbf{13.26\%}} &  {\ul \textbf{33.3\%}} &  {\ul \textbf{45.13\%}} &  {\ul \textbf{60.47\%}} & {\ul \textbf{2.26\%}} & {\ul \textbf{7.29\%}} & {\ul \textbf{11.15\%}} & {\ul \textbf{18.94\%}} \\
C-CLIP-desc-mixer-200k & 38.6\% & 71\% & 81.7\% & 92.89\% & 13.3\% & 31.76\% & 43.36\% & 58.56\% & 2.18\% & 7.07\% & 10.82\% & 17.8\% \\
%C-CLIP-mixer-big & 23.3\% & 45.09\% & 58.2\% & 75.29\% & 9.6\% & 22\% & 30.7\% & 46.6\% & -\% & -\% & -\% & -\% \\
\end{tabular}%
}
\caption{Retrieval accuracy results for all models trained on the two commentative data sets (top: Telegram, bottom: Twitter). The best performing model on the telegram data was C-CLIP-tele-2m, perhaps unsurprisingly. Increasing the size of the training set appears to yield increasing results. The highest accuracy for the twitter data set was actually a heterogenous mixture of twitter and telegram data. Mixing twitter and descriptive data also did well, but not as well as supplementing with commentative data.}
\label{tab:tele_results}
\end{table*}

% \usepackage{multirow}
% \usepackage{graphicx}
% \usepackage[normalem]{ulem}
% \usepackage{multirow}
% \usepackage{graphicx}
% \usepackage[normalem]{ulem}

%Initial experiments were attempts to answer the following questions:

%\begin{enumerate}[noitemsep,nolistsep]
%    \item Does the DCG exist in available CLIP models?
%    \item If it exists, what is the magnitude of difference on downstream tasks?
%    \item Is it persistent across different languages?
%    \item Is it persistent across different social media sites?
%\end{enumerate}

With the four datasets chosen for testing the initial hypotheses as outlined in the Methodology the results in Table~\ref{tab:baselines} are produced. They show a clear degradation in performance for baseline CLIP models when they move from a descriptive task to a commentative one. Additionally, these problems exist both across languages and social media sites, demonstrating that the DCG describes something fundamentally present in how social media is used differently when compared to the types of image-text pairs that CLIP models are commonly trained on.

The degradation in results is larger than one might expect, with both multilingual baseline models dropping 61.78\% and 65.61\% percentage points at a recall accuracy @10 for the Russian Telegram data and 54.31\% and 81.04\% percentage points on the Brazilian Tweet data set when compared to XM3600. All results are the average accuracy over 10 trials, with a population size of 1,000 image-text pairs, expect for XM3600, which having only 3600 pairs means that at a population size of 1,000 produces only 3 test splits. The standard deviation across all trials on all data sets on all experiments averaged 2.34\% at recall 10, making the difference between 3 trials and 10 trials negligible when considering differences of approximately 30 percentage points.

The first three rows in Table~\ref{tab:baselines} demonstrates rather conclusively that the DCG results in decreased accuracy of models trained purely on descriptive data when applied to commentative tasks, estimates the magnitude of its effect, and shows that it persists across languages and (at least two) social media sites.

These baselines show that there is much need for improvement on downstream tasks vital to operators in OSINT spaces.

Our best results on the two commentative data sets are shown in the bottom two rows. Training on the specific data set massively increases performance. While perhaps not particularly surprising, it's  important to show that the commentative image-text difference is possible to learn. Our best result on the Russian Telegram data is 36.51 percentage points higher (again @10) and for the Brazilian data, 28.2 percentage points higher than the best baseline.

Quantitative results give only one side of the story, the qualitative side is also important. The retrieval accuracy @5 is only 57.25\% on the Russian Telegram data set. However if one considers Figure~\ref{fig:quali} it can be seen that the other 4 results that are not a direct match are still highly relevant to the query text, in both sets of retrieval results. The top results contain 4 images of monuments and one of a soviet symbol, all of which are mentioned in the query text. The second set of results is for a piece of text about a petition that is being sent to President Zelensky to review. 4 out of the 5 returned results are about various petitions that have been filed since the invasion began. It seems reasonable that even if the directly matching image is not returned, OSINT operators would find useful information in non-match results. If a journalist or reporter were writing a piece on junk petitions filed during the on-going invasion, having a tool to find all posts related to petitions seems a useful tool.  The achieved accuracy scores are state-of-the-art, but we contend that the results are even more useful than these scores imply, especially to real-world operators in OSINT.

We also report the "backwards" accuracy of the new models, I.E. how the models trained on commentative data perform on descriptive tasks. In this we can see the DCG happening in reverse, demonstrating that there is indeed a gap. Training on descriptive data only means low accuracy when tested on commentative data and training on commentative data only results in low accuracy when tested on descriptive data. The DCG appears to be a two way street.

In addition to experiments exploring the nature of the DCG, we also report results on a variety of training variations used on our models. Table~\ref{tab:tele_results} shows the results of all models on the Russian Telegram data set and the Brazilian Twitter data set. The first row shows the baseline with the highest score on the data set for reference. For Russian Telegram the highest accuracy was achieved by the model that was given the most training data, in this case 2 million image-text pairs. All training pairs were filtered to ensure that the associated text had at least 5 "words" (quotations being used on words as there were many emojis in the data, potentially another interesting avenue to explore).

Varying the amount of training data was not the only vector through which accuracy could be increased. The selection of training data is an active area of research and experiments were performed to see to what extent mixing and matching training data would have on downstream accuracy. Due to the relatively small size of the Brazilian Twitter data set (to successfully run 10 distinct trials of the 3 populations you need 111,000 test pairs) the maximum number of Brazilian twitter pairs the model could be trained on was limited to 100,000. This resulted in an accuracy of only 31\% (pop=1000, recall=10). While this result is nearly twice the baseline accuracy, it seemed likely that supplementing the training data with other image-caption pairs could lead to an increase. Four different models were tested: one with 100,000 Brazilian tweets and 100,000 Russian telegram posts, a training set with 100,000 Brazilian tweets and 100,000 descriptive pairs, a set with 100,000 Brazilian tweets, 100,000 descriptive pairs, and 100,000 telegram pairs, and one with 100,000 Brazilian tweets, 100,000 descriptive pairs, and 2m telegram pairs. Out of these four, the model that was trained on an equal split of Brazilian Twitter and Russian Telegram data performed the best, increasing to 45.13\%. This was closely followed by the other two models that had even data splits, the one trained on 100,000 pairs of both descriptive and Brazilian data and the model trained on Brazilian, Russian, and descriptive data. Interestingly, the model that saw the most data, but had a large skew in the data away from the Brazilian twitter data set performed the worse of the four (though again still better than the baseline). This seems to imply that while descriptive and commentative data are two distinct categories, there exist sub categories of the two that are more specific to certain languages and/or sites. Supplementing with data of a similar type (commentative/descriptive) may help but you can't entirely replace the in-task data with data of the same class.

The model that was trained on Brazilian twitter, Russian Telegram, and descriptive data was tested to see how well its accuracy extended across all four data sets. The goal is a model that achieves universally, uniformly, high accuracies on all tasks. Unfortunately this doesn't seem to be the case. As can be seen in Table~\ref{tab:baselines}, the C-CLIP-mixer-300k (last line) doesn't achieve particularly high accuracy results. While one could argue that the differences in amount of training data could potentially make up for this, the accuracy on specifically the Russian Telegram data set is much lower than its counterpart model that was trained on the Russian data only. Compared to the C-CLIP-tele-100k model seen in Table~\ref{tab:tele_results}, the C-CLIP-mixer-300k got 12.25 percentage points lower retrieval accuracy (pop=1000, recall=10). This is rather surprising, as supplementing with more data increased accuracy on the Brazilian Twitter data set, but seems to lower it on the Russian Telegram data set. As hypothesized above, there seem to be sub-categories in the overarching classes of descriptive and commentative that contain their own quirks. While our models achieve state-of-the-art accuracy scores across the board, further research into this area is greatly needed.

Brief experiments were also run to explore the extensibility of models on data sets they were not trained on. The results can be seen in Table~\ref{tab:extens}. Shown are the differences between the cosine similarity of a true pair and the average of all false matches, for 1000 image-text pairs. Unsurprisingly we see an increase in this difference as C-CLIP-tele is trained on increasingly more data and this is reflected in increasing accuracy scores as can be seen in Table~\ref{tab:tele_results}. What is interesting is the growth (or lack-there-of) in the differences on the MSCOCO and Brazilian Twitter data sets. If types commentative data were truly indistinguishable from themselves then we could expect the model to improve its results on the Brazilian data set at a similar pace as the Telegram data set. However we instead see no such increase. This seems to support the intuition stated above that there exist sub-sections of data falling under the commentative super-class. This is reflected qualitatively in the TSNE plots shown in Figure~\ref{fig:dcg} but required the additional quantitative justification provided here. Therefore it seems that if one wishes to have an accurate C-CLIP model the best data to train on is data specific to their task.

If one has access to a limited amount of data for that task, supplementation of data does help. These results can be seen in the bottom three rows of Table~\ref{tab:extens}. Training on a mixture of balanced commentative data yields the best result, and is actually the highest performing model on the Brazilian Twitter data set (shown in ~\ref{tab:baselines}). Supplementing this model further with descriptive data actually decreases the difference between positive and negative similarities. With the Twitter models we also see the trend of a lack of increase in the models performance on other commentative data sets, further supporting the notion that there exist subtle differences in commentative data. The cause of these differences are unknown but could be as far ranging as language specific to the event or region, or just different emoji usage by different cultures. Further work in this area is required. 

The DCG implies that there should not be an increase on the MSCOCO data set when training on commentative data, but we instead see a small increase in the differences on the MSCOCO data set, even when a model is not trained on descriptive data. We believe that the underlying models already having a strong baseline in descriptive tasks is able to slowly appear as the projection layers get more and more finely tuned with increasingly large amounts of commentative data. However as the mixer-300k and desc-mixer-200k models show, the best way to increase the difference is simply to have descriptive data present in the training data.

% \usepackage{multirow}
% \usepackage[normalem]{ulem}
% \usepackage{multirow}
% \usepackage[normalem]{ulem}
% \usepackage{multirow}
% \usepackage{graphicx}
% \usepackage[normalem]{ulem}
% \usepackage{multirow}
% \usepackage{graphicx}
% \usepackage[normalem]{ulem}
% \usepackage{multirow}
% \usepackage{graphicx}
% \usepackage[normalem]{ulem}

% \usepackage{graphicx}
\begin{table}[]
\centering
\begin{tabular}{c|ccc}
C-CLIP Model & MSCOCO & Telegram & Twitter \\ \hline
tele-10k & 0.1172 & 0.2387 & 0.1081 \\
tele-100k & 0.2578 & 0.3422 & 0.1359 \\
tele-1m & 0.2589 & 0.3847 & 0.1147 \\
tele-2m & 0.2056 & 0.4297 & 0.0906 \\ \hline
twit-10k & 0.0964 & 0.0764 & 0.1910 \\
twit-100k & 0.0866 & 0.0491 & 0.2499 \\
desc-mixer-200k & 0.4826 & 0.0800 & 0.2981 \\
mixer-200k & 0.1723 & 0.2771 & 0.3438 \\
mixer-300k & 0.4763 & 0.2468 & 0.3083
\end{tabular}
\caption{The average difference between the cosine similarity of a correctly matching pair and the average of all incorrect pairs for a population size of 1000. The ideal model has a large difference between the similarities of a positive pair and a negative pair and displays this difference across all three data sets.}
\label{tab:extens}
\end{table}

\section{Conclusions}

By training CLIP models on commentative text rather than only descriptive text we can improve the accuracy of these models in downstream retrieval tasks on social media data. While we report state-of-the-art retrieval accuracies when compared to baseline multilingual clip models, the usefulness of the results is better than the accuracy implies.
Allowing non-technical people to explore large collections of unsorted and unlabelled data and discover semantically similar results with higher speed and accuracy is key to improving OSINT operations. Figure~\ref{fig:quali} shows that results that are not an exact match, as determined by the actual pairing from social media, are still highly relevant and further integration of C-CLIP models into pre-existing downstream tech is an exciting prospect.

\textbf{Limitations and Future Work.} When training models the limitations are of course primarily data related. Training data for the C-CLIP models was randomly selected from the data set, and a better curation method would likely lead to better results. Additionally the creation of more social media data sets in a variety of languages is crucial to improving understanding in languages outside of English. This paper merely trains the models and measures baselines, much future work is yet to be done in integrating these models into OSINT tools in order to aid those involved.

%\textbf{Acknowledgements.}
%This material is based on research sponsored %by the Defense Advanced Research Projects %Agency (DARPA) and the Air Force Research %Laboratory (AFRL) under AFRL agreement number %FA8750-16-2-0173. The U.S. Government is %authorized to reproduce and distribute %reprints for Governmental purposes %notwithstanding any copyright notation %thereon. The views and conclusions contained %herein are those of the authors and should %not be interpreted as necessarily %representing the official policies or %endorsements, either expressed or implied, of %DARPA, AFRL, or the U.S. Government.

%%%%%%%%% REFERENCES
{\small
\bibliographystyle{ieee_fullname}
\bibliography{egbib}
}

\begin{figure}[t!]
\includegraphics[width=1.96\linewidth]{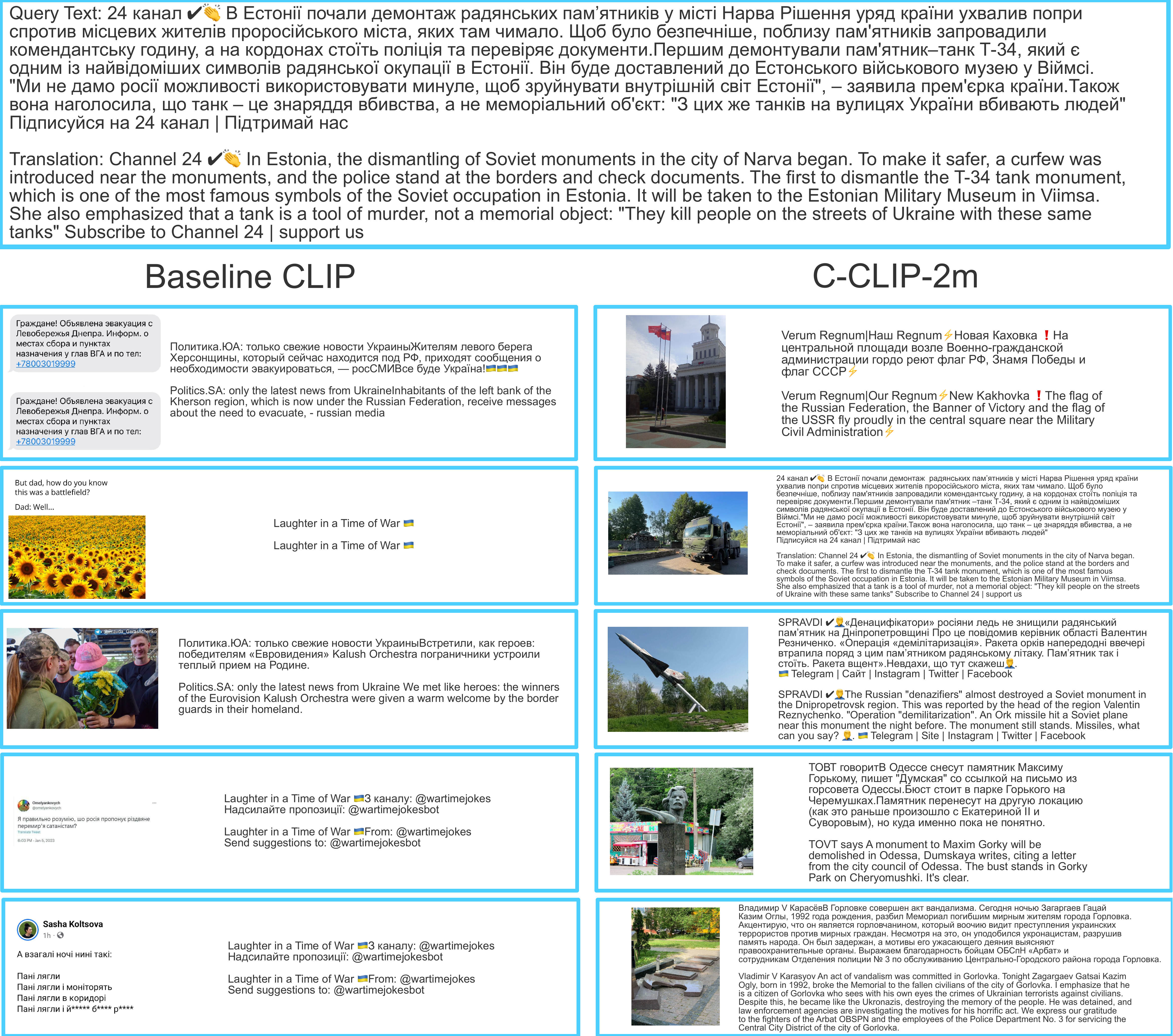} % Reduce the figure size so that it is slightly narrower than the column.
\vspace{-2mm}
\caption{The qualitative results with the original, untranslated captions that were posted with the image. Google translate was used for the translation, so slight inconsistencies may exist. Hence the need to include captions in the original.}
\vspace{-2mm}
\label{fig:quali}
\end{figure}

%\end{document}

\end{document}